# A Multimodal Autonomic Sensing Framework for Objective Assessment of Patient Responses to Dental Pulp Stimulation

Youngsun Kong, Yubin Choi, Dongjin Song, Dong-Guk Shin, I-Ping Chen, and Ki Chon

***Abstract*—Patient responses to dental pulp testing, ranging from no sensation to intense pain, provide important information for assessing pulp status in endodontic diagnosis. However, pain is a subjective sensory and emotional experience that varies considerably across individuals and can be difficult to communicate. We investigated whether complementary autonomic signals could support objective assessment of responses during dental examination. Forty-nine patients underwent cold pulp testing, yielding no-response, mild-response, and intense-response conditions. The framework integrated ECG-derived skin nerve activity (SKNA) and R–R intervals (RRI), together with electrodermal activity (EDA), using temporal convolutional network encoders with attention-based mid-level fusion. Individual baseline signals and subject-level covariates, including anxiety scores and biological sex, were also incorporated. The framework achieved 80.2% balanced accuracy, 75.2% sensitivity, and 85.2% specificity for binary classification of no response versus mild or intense response. For three-class classification, it achieved 60.0% balanced accuracy and a 58.8% macro-averaged F1 score. Ablation and attention-weight analyses indicated that EDA contributed most strongly to model performance, followed by RRI, while SKNA improved balanced accuracy by approximately five percentage points. Age was significantly associated with model performance. These findings support the feasibility of multimodal autonomic sensing for objective, non-invasive assessment of responses to dental pulp stimulation.**

## I. INTRODUCTION

Dental caries is the most prevalent health condition worldwide and its burden continues to rise with population growth [1]. Bacteria from caries can invade the root canal system, leading to infection of the dental pulp and periapical tissues [2]. This process can result in hypersensitivity, severe pain, and, in rare cases, life-threatening complications [3]. In clinical practice, dental pain is primarily assessed using subjective measures such as the visual analog scale (VAS), where patients rate pain intensity on a scale from 0 (no pain) to 10 (worst imaginable pain) [4]. However, the lack of objective and quantitative measures for dental pain presents significant challenges, including misdiagnosis, difficulty assessing patients with limited communication abilities (e.g., young children), and potentially inappropriate analgesic prescribing. Therefore, there is a critical need for objective, physiology-based markers to enable quantitative assessment of dental pain.

Heart rate (HR) and heart rate variability (HRV) are among the most commonly used non-invasive measures of autonomic nervous system (ANS) activity. Derived from cardiovascular recordings, such as electrocardiogram (ECG) and photoplethysmogram (PPG), these metrics reflect autonomic modulation through variability in beat-to-beat intervals (e.g., R–R intervals, RRI). Due to their accessibility, early studies of dental pain often utilized HR and HRV to characterize autonomic responses [5], [6], [7]. However, interpretation of these measures require caution, as they do not provide direct indices of sympathetic activity and are strongly influenced by parasympathetic regulation and overall sympathovagal balance. Moreover, HRV analyses are often conducted over relatively long time windows (e.g., baseline, anesthesia, surgical procedure, and recovery) [6], [8], limiting their sensitivity for capturing rapid and transient autonomic responses elicited by acute dental pain stimuli.

In contrast, electrodermal activity (EDA) and skin nerve activity (SKNA) enable assessment of rapid autonomic responses to discrete dental stimuli, such as cold or electric pulp testing (EPT). EDA reflects sudomotor activity mediated by sympathetic cholinergic pathways (acetylcholine) [9], whereas SKNA provides a more direct measure of sympathetic nerve activity, which can be extracted as high-frequency components from ECG recordings [10]. Compared to HR- and HRV-based measures, these modalities offer greater temporal sensitivity for detecting acute autonomic responses, as cardiovascular changes

This research was supported by the National Institute of Dental & Craniofacial Research of the National Institutes of Health under Award Number F32DE033566. (Corresponding author: Youngsun Kong, co-corresponding author: I-Ping Chen).

Youngsun Kong is with the Department of Biomedical Engineering, University of Connecticut, Storrs, CT 06269 USA (e-mail: youngsun.kong.phd@gmail.com).

Yubin Choi was with the Department of Endodontics, University of Connecticut Health, Farmington, CT 06030 USA (e-mail: yuchoi@uchc.edu).

Dongjin Song is with the School of Computing, University of Connecticut, Storrs, CT 06269 USA (e-mail: dongjin.song@uconn.edu).

Dong-Guk Shin is with the School of Computing, University of Connecticut, Storrs, CT 06269 USA (e-mail: dong.shin@uconn.edu).

I-Ping Chen is with the Department of Endodontics, University of Connecticut Health, Farmington, CT 06030 USA (e-mail: ipchen@uchc.edu).

Ki Chon is with the Department of Biomedical Engineering, University of Connecticut, Storrs, CT 06269 USA (e-mail: ki.chon@uconn.edu).

are mediated through relatively slower adrenergic mechanisms [11].

Based on prior studies, we hypothesize that integrating complementary autonomic information from R-R intervals (RRI), EDA, and SKNA could enable a multimodal autonomic sensing framework for objectively assessing patient responses to dental pulp stimulation. Notably, both RRI and SKNA can be derived from surface ECG recordings, enabling streamlined data acquisition with a single sensor setup. While EDA and SKNA both reflect sympathetic activity, they arise from distinct physiological pathways, providing complementary information. We therefore propose a multimodal framework for objective assessment of dental pain responses using EDA, SKNA, and RRI. The key contributions of this work are as follows: 1) development of multimodal deep learning framework with attention-based mid-level fusion to capture temporal autonomic dynamics; 2) a physiologically-informed design, including selection of window lengths and preprocessing strategies based on autonomic signal characteristics; and 3) design of a classification framework aligned with clinical diagnostic workflows, focusing on objective assessment during clinician-administered testing rather than relying solely on subjective pain reporting.

## II. Related Work

### A. Automatic Assessment of Dental Pain and Pulp Responses

Early efforts toward automatic assessment of responses to dental pulp stimulation were already multimodal. In 2005, [12] combined electromyographic (EMG) activity from the anterior belly of the digastric muscle, finger movement, and vocal responses during EPT. Each signal was processed independently using predefined thresholds, and detection of a pulp-evoked response in any modality triggered termination of the electrical stimulus, with the goal of minimizing unnecessary painful stimulation. Although this approach did not employ machine learning or learned multimodal representations, it represents an early example of automatic multimodal detection of pulp-evoked responses.

Following this early work, there has been a relative paucity of studies specifically addressing automatic assessment of responses to dental pulp stimulation. In contrast, the broader field of automatic pain recognition (non-dental pain) has expanded substantially over the same period. In particular, the BioVid Heat Pain Database, introduced in 2013 [13], provided experimentally induced heat-pain responses together with physiological and facial recordings and subsequently became a widely used databank for automatic pain-recognition research. Broader developments in automatic pain recognition and multimodal sensing are discussed in Section II-B.

Subsequent work on automatic dental pain recognition was reported by Teichmann *et al.* [14], who investigated pain occurring during clinical dental examinations. They analyzed 6-s segments of ECG, PPG, and thoracic respiratory effort signals collected from 20 patients and evaluated several machine-learning classifiers, including multilayer perceptron (MLP), support vector machine (SVM), k-nearest neighbors, and random forest models (RF). The random forest classifier achieved a sensitivity of 87% and a specificity of 63% for distinguishing pain from no-pain periods during cold testing and periodontal examination. This study demonstrated the feasibility of identifying dental pain events from autonomic and cardiorespiratory signals in a clinical setting. However, the framework was designed for general pain-event detection across different dental examination procedures rather than specifically characterizing the physiological response to dental pulp stimulation.

One of our recent studies investigated an EDA-based approach for automatic dental pain assessment [15]. In that study, EDA was collected from 51 patients undergoing EPT, and logistic regression (LR), SVM, RF, and MLP classifiers were evaluated using time-varying and phasic (i.e., rapid response component)-derived EDA features. The MLP achieved a sensitivity of 76% and a specificity of 87% for detecting clinically meaningful pain, defined as a VAS score of ≥4 on a 0–10 scale. However, EPT is primarily intended to elicit and assess pulpal sensibility rather than to reproduce clinically relevant dental pain. Accordingly, this framework focused on identifying the presence of meaningful pain during EPT rather than characterizing the physiological response to a standardized painful pulp stimulus, such as cold testing.

Other sensing modalities have also been explored for dental pain assessment. Earlier work incorporated vocal responses and behavioral movements during EPT [12]. Additionally, Stillhart *et al.* investigated automated facial-expression analysis in patients presenting with dental pain before and after treatment [16]. Although facial behavior has been widely investigated in the broader automatic pain-recognition literature, its application during dental pulp testing may be more challenging because facial regions can be partially occluded by the clinician, hands, or dental instruments during the examination.

Overall, objective and automatic assessment of dental pulp responses has remained largely centered on physiological and autonomic sensing. Existing approaches have primarily focused on detecting the presence of a response, distinguishing pain from no pain, or classifying pain according to self-reported intensity measures such as the VAS. In contrast, limited work has investigated whether multimodal autonomic responses can distinguish clinically defined response patterns during dental pulp testing. The present study addresses this gap by integrating multiple autonomic sensing modalities to characterize distinct patient responses to cold pulp stimulation, ranging from no sensation to mild sensation and intense pain.

### B. Automatic Pain Recognition Using Machine Learning

Beyond dental applications, automatic pain recognition has been extensively investigated using physiological, behavioral, and multimodal sensing. Many existing approaches have been developed around established benchmark datasets, including the BioVid Heat Pain Database [13], SenseEmotion [17], X-ITE Pain Database [18], PainMonit [19], and AI4PAIN [20]. These datasets provide combinations of physiological and behavioral modalities collected primarily during experimentally induced heat and/or electrical pain and have supported the development and evaluation of a wide range of automatic pain-recognition methods [21], [22].

In those public datasets, physiological modalities commonly include ECG, PPG, EMG, respiration, and EDA, while behavioral indicators include facial expressions captured through video, body movements, and vocal responses. Although individual modalities have been investigated independently for automatic pain recognition, many studies have focused on multimodal approaches that combine complementary information across different sensing modalities. Comparative and ablation analyses in these studies have often demonstrated improved pain-recognition performance when multiple modalities are combined rather than used individually [21], [22].

Earlier automatic pain-recognition studies primarily relied on handcrafted features and conventional machine-learning classifiers such as SVM and RF. Werner *et al.* combined physiological signals, including ECG, EMG, and EDA, with facial expressions and head movements for automatic pain recognition using the BioVid dataset [23]. Using RF, they achieved accuracies of 80.6% and 77.8% for personalized and generalized models, respectively, in distinguishing baseline from the highest pain level. Building on the same dataset, Kächele *et al.* achieved accuracies of 82.7% and 83.1% using early- and late-fusion strategies, respectively, for distinguishing baseline from the highest pain level [24].

With the increasing adoption of deep learning, convolutional neural network (CNN)- and recurrent neural network (RNN)-based approaches have been used to learn modality-specific representations and temporal dynamics associated with pain. For example, Thiam *et al.* investigated deep CNN-based architectures for learning pain-related representations from physiological signals in the BioVid dataset [25]. Their approach achieved an accuracy of 79% by combining EMG, ECG, and EDA using a late-fusion strategy for distinguishing baseline from the highest pain level. Wang *et al.* subsequently employed a bidirectional long short-term memory network to model temporal dynamics across EDA, ECG, and EMG signals and combined the learned representations with handcrafted features using a hybrid RNN–artificial neural network (ANN) architecture, achieving an accuracy of 83.3% for the same classification task [26].

More recent studies have incorporated attention mechanisms and Transformer-based architectures to model interactions among complementary sensing modalities. Thiam *et al.* proposed an attention-enhanced multimodal architecture that learned modality-specific representations from EDA, ECG, EMG, and respiration and adaptively weighted the resulting physiological features, achieving an accuracy of 84.3% for distinguishing baseline from the highest pain level on the BioVid dataset [27]. Gkikas *et al.* employed Transformer-based temporal modeling to integrate facial-video and heart-rate representations for multimodal acute-pain assessment, achieving an accuracy of 82.7% for the same binary classification task [28]. More recently, Farmani *et al.* proposed a CrossMod-Transformer that separately modeled EDA and ECG representations before learning inter-modal dependencies between the two physiological signals, achieving an accuracy of 87.5% on the BioVid dataset [29].

In summary, recent automatic pain-recognition research has increasingly adopted attention-based and cross-modal fusion approaches to integrate complementary representations learned from multiple sensing modalities. Building on these advances, we developed a multimodal autonomic framework specifically tailored to objective assessment of patient responses to dental pulp stimulation. The framework was designed with several clinical considerations in mind, including the practicality of sensing modalities in the dental environment, where facial sensing can be limited by occlusion; the influence of individual factors, such as anxiety and biological sex, on autonomic responses; and a prediction target that extends beyond pain intensity alone. Specifically, the present task aims to distinguish a range of clinically relevant responses to pulp stimulation, including no response, mild sensation, and intense pain. Accordingly, the proposed framework employs modality-specific CNN encoders followed by attention-based fusion to integrate complementary autonomic representations for this clinical assessment task.

## III. Methods

### *A. Participants*

Participants were recruited from patients undergoing treatment at endodontic clinics at UConn Health. Eligible participants were those requiring root canal treatment for one tooth and having two nearby healthy teeth. Exclusion criteria included individuals under 18 years of age, pregnant women, teeth with porcelain crowns, individuals taking medications with anticholinergic side effects (which may affect skin conductance), those with a prior history of sympathectomy, and patients diagnosed with Raynaud's syndrome.

A total of 50 participants were enrolled in the study; data from one participant were excluded due to wire disconnection, resulting in a final cohort of 49 participants (21 males and 28 females; age range: 21–68 years). Informed consent and Health Insurance Portability and Accountability Act (HIPAA) authorization were obtained prior to the experiments. All procedures involving human subjects were conducted in accordance with the Institutional Review Board of the University of Connecticut Health (IRB protocol 20-043-1).

### *B. Dental stimulus — cold testing*

Cold test was performed using a refrigerant spray, Endo-ice (Coltene, Altstätten, Switzerland), which can reach temperatures as low as −26.2 °C. The dentist applied the refrigerant spray to a cotton pellet, which was then briefly placed on the target tooth. Participants were instructed to raise their hand to indicate the onset of stimulus perception. Following each stimulus, participants reported perceived intensity using a 0–10 VAS. Subsequently, the dentist provided a clinical adjudication of the response as one of the following: no response (negative), mild response (+), or intense response (++). Fig. 1 shows the distribution of VAS scores across cold testing outcomes. Depending on pulp status, responses may range from mild to intense sensations in vital or inflamed teeth, whereas no response is typically observed with necrotic teeth [30].

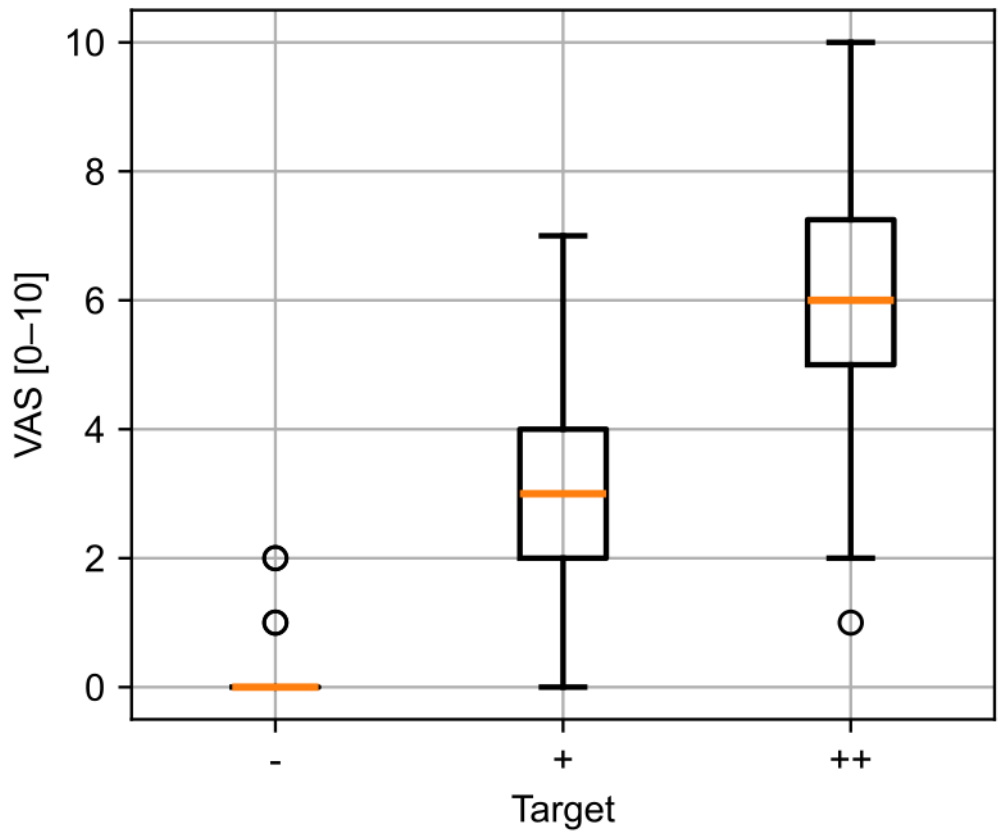

**Fig. 1.** VAS distribution across cold testing outcomes. -: negative responses, +: mild responses, ++: intense responses.

A minimum of three teeth were tested per participant, including one or more affected teeth and two nearby healthy teeth. A sham test, meaning a clean, unsprayed cotton pellet was applied to the tooth, was included. The interstimulus interval was set to at least 20 seconds to allow perceptual recovery between trials.

The proposed framework targets two classification tasks: 1) negative versus positive responses, and 2) negative/mild versus intense responses. The first task aims to differentiate tooth vitality, while the second focuses on identifying pulpitis, which is typically associated with heightened pain responses.

*C. Data Acquisition*

EDA and ECG signals were acquired using a clinical-grade Galvanic Skin Response (GSR) Amp device and a BioAmp system, respectively, with a PowerLab amplifier (ADInstruments, Sydney, Australia). EDA was recorded from the index and middle fingers of the non-dominant hand using reusable stainless-steel electrodes, as the dominant hand was used to indicate stimulus onset. ECG signals were collected using Ag/AgCl electrodes placed on the inner wrist of the non-dominant hand and the inner ankle of the contralateral side, with a reference electrode positioned on the opposite ankle. Signals were sampled at either 10 kHz or 4 kHz using LabChart 8 software (ADInstruments, Sydney, Australia). Prior to electrode placement, the skin was cleaned with 70% isopropyl alcohol wipes to ensure signal quality. Following instrumentation, participants rested in a seated position for at least two minutes to allow physiological stabilization before undergoing the cold stimulation protocol. Fig. 2 illustrates the electrode placement locations.

*D. Feature extraction*

1) **Electrodermal activity (EDA)**

EDA consists of two primary components: phasic and tonic. The phasic component captures rapid, transient changes in skin conductance associated with stimulus-evoked responses, whereas the tonic component reflects slower variations corresponding to baseline sympathetic tone. Although both components are responsive to pain and sympathetic activation [31], recent studies have primarily leveraged phasic or higher-frequency components for machine learning applications due to their stronger temporal association with stimulus-driven dynamics. In particular, derivatives of the phasic EDA (PhEDA) component have demonstrated strong discriminative power in pain classification tasks [32]. Additionally, time–frequency representations such as spectrograms and phasic drivers (PhdEDA) have been successfully incorporated into CNN–based models [33].

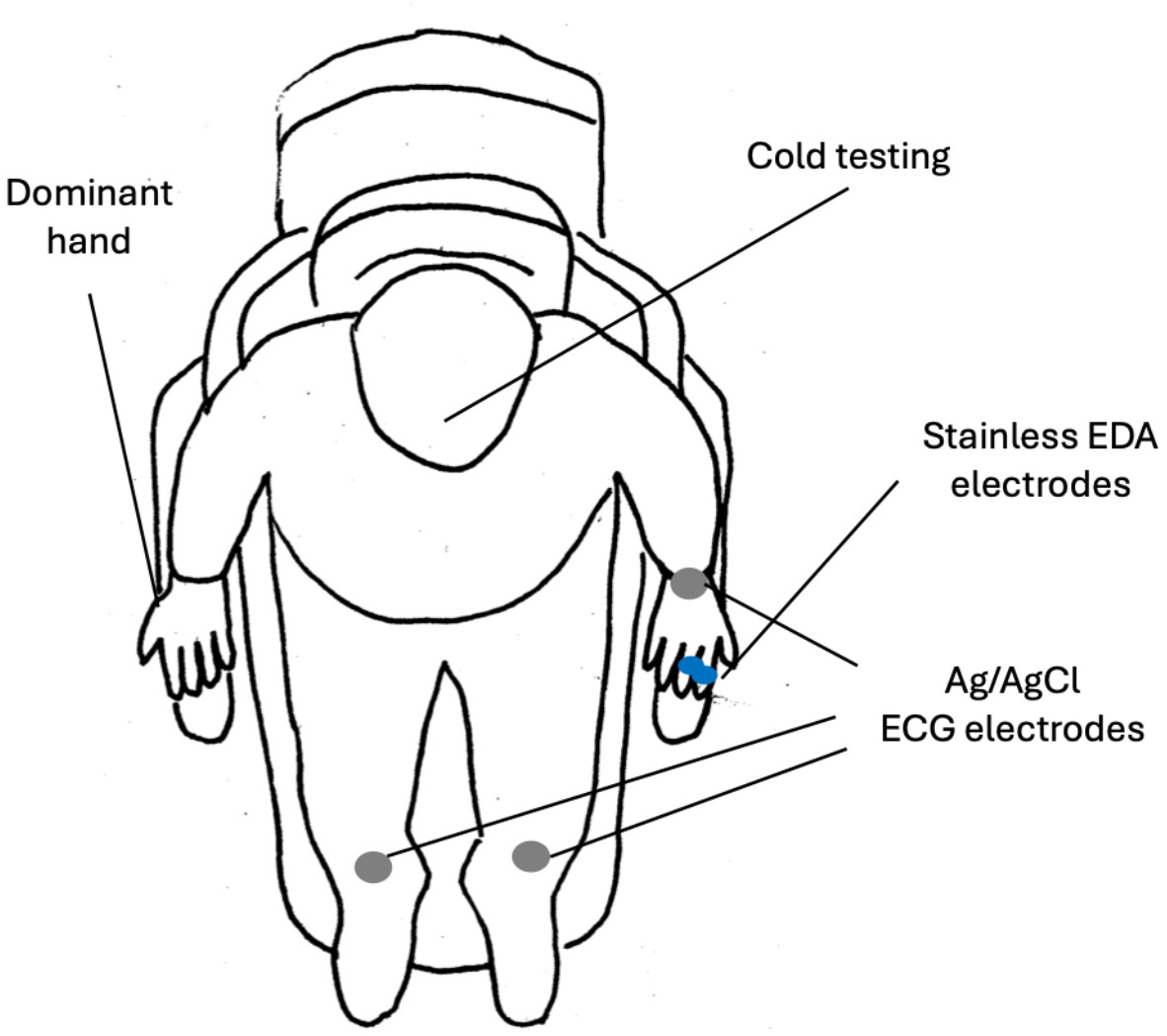

**Fig. 2.** Electrode placement locations.

EDA signals were first resampled to 4 Hz and highpass filtered at 0.01 Hz to remove slow baseline drift. A short-time Fourier transform (STFT) was then applied using a Hann window with a 2 s window length and 1.5 s overlap, with $n_{FFT}$ =16, yielding 8 frequency coefficients with a 2 Hz sampling frequency. Using the cvxEDA method [34], EDA was decomposed into phasic and tonic components, and the corresponding phasic driver was estimated. To align temporal resolution with the spectrogram, phasic EDA component and its phasic driver were resampled to 2 Hz. The first-order derivative of the phasic component (dPhEDA) was computed using a Savitzky–Golay filter (window length = 11 samples, polynomial order = 2) [35]. In total, 11 time-series features were extracted from the EDA signals.

2) **R-R intervals (RRI)**

Although HRV indices are commonly used for autonomic assessment, they typically require data lengths on the order of one minute for reliable assessment, limiting their applicability for rapid, stimulus-driven analysis [36]. Instead of extracting conventional HRV metrics, R–R intervals (RRI) and their derivative features were directly used for assessment. Prior studies have demonstrated the effectiveness of directly using RRI in CNN–based models [37], [38], [39].

R-peaks were detected from ECG signals using Complete Ensemble Empirical Mode Decomposition (CEEMD) [40] after resampling the signals to 1 kHz. Detected peaks were then manually reviewed and corrected to ensure accuracy. Because R-peak occurrences are inherently irregular in time, the resulting R–R intervals were resampled to 4 Hz using linear interpolation. The first-order derivative of the RRI signal (dRRI) was computed using a Savitzky–Golay filter (window length = 21 samples, polynomial order = 2) [35]. In total, two time-series features were derived from the RRI signals.

3) **Skin nerve activity (SKNA)**

SKNA is commonly quantified using integrated SKNA derived from the neuECG technique [10]. A more recent approach, known as time-varying SKNA (TVSKNA), has demonstrated enhanced sensitivity, reproducibility, and accuracy for detecting SNS activity [41]. Compared with EDA and RRI, SKNA contains higher-frequency components and is acquired at a higher sampling rate, enabling characterization of rapid sympathetic dynamics. Instead of directly using TVSKNA and its time-series derivatives as inputs, as was done for EDA and RRI, descriptive and complexity-based features were extracted from SKNA using 5 s windows with 4.5 s overlap, yielding in time-resolved feature sequences with a 2 Hz sampling frequency. This design choice was made to balance model complexity across modalities given the relatively limited data size. In particular, complexity features have been shown to provide complementary information to amplitude-based measures and are effective for capturing short-term sympathetic activation across various SNS-inducing tasks [42].

ECG signals were first downsampled to 4 kHz and high-pass filtered at 150 Hz to isolate high-frequency components associated with SKNA. Narrowband interference was identified using power spectral density (PSD) analysis on a per-participant basis, with visual inspection used to confirm distinct spectral peaks. Frequencies consistently present in both baseline and task conditions were treated as condition-independent artifacts and removed using second-order IIR notch filters (Q = 30).

The computation of TVSKNA consists of three steps. First, the filtered signals were decomposed into 12 frequency bands using variable frequency complex demodulation (VFCDM) [43] to obtain a time–frequency spectrum (TFS), followed by signal reconstruction via summation of the TFS components within the 480–1120 Hz range. Second, the instantaneous amplitude was estimated using the Hilbert transform. Finally, the resulting signal was smoothed using a moving average filter with a 100 ms window. Additional methodological details are provided in our previous work [41].

From the resulting TVSKNA time series, the following features were extracted: mean, maximum, minimum, mean absolute deviation, standard deviation, Hjorth mobility and complexity [44], detrended fluctuation analysis (DFA) [45], Katz fractal dimension (KFD) [46], sample entropy (SampEn, m = 2, r = 0.15 × standard deviation) [47], and approximate entropy (ApEn, m = 2, r = 0.15 × standard deviation) [48]. The parameters for entropy measures were selected based on prior analysis [42]. A total of 11 time-series features were computed for SKNA.

4) **Subject-level covariates (auxiliary features)**

In addition to ANS features, two subject-level covariates were incorporated as auxiliary inputs. Among the available demographic and clinical variables, biological sex and dental anxiety scores (DAS) were selected based on their balanced distribution within the study cohort and their known influence on autonomic responses, as reported in prior studies [49], [50], [51].

The DAS is a four-item questionnaire, with each item scored on a five-point scale (1–5), yielding a total score range of 4–20. Clinically, DAS scores are interpreted as follows: moderate anxiety (9–12), high anxiety (13–14), and severe anxiety (15–20) [52].

## *E. Proposed Framework*

Fig. 3 illustrates the overall architecture of the proposed multimodal framework, which integrates EDA, RRI, and SKNA. Domain-specific time-series features are first extracted from each autonomic modality, followed by CNN–based embedding to capture temporal patterns. The learned representations are then integrated using an attention-based fusion mechanism for dental stimulus classification. In addition, subject-level covariates, including biological sex and anxiety scores, are incorporated into the framework to enhance model performance.

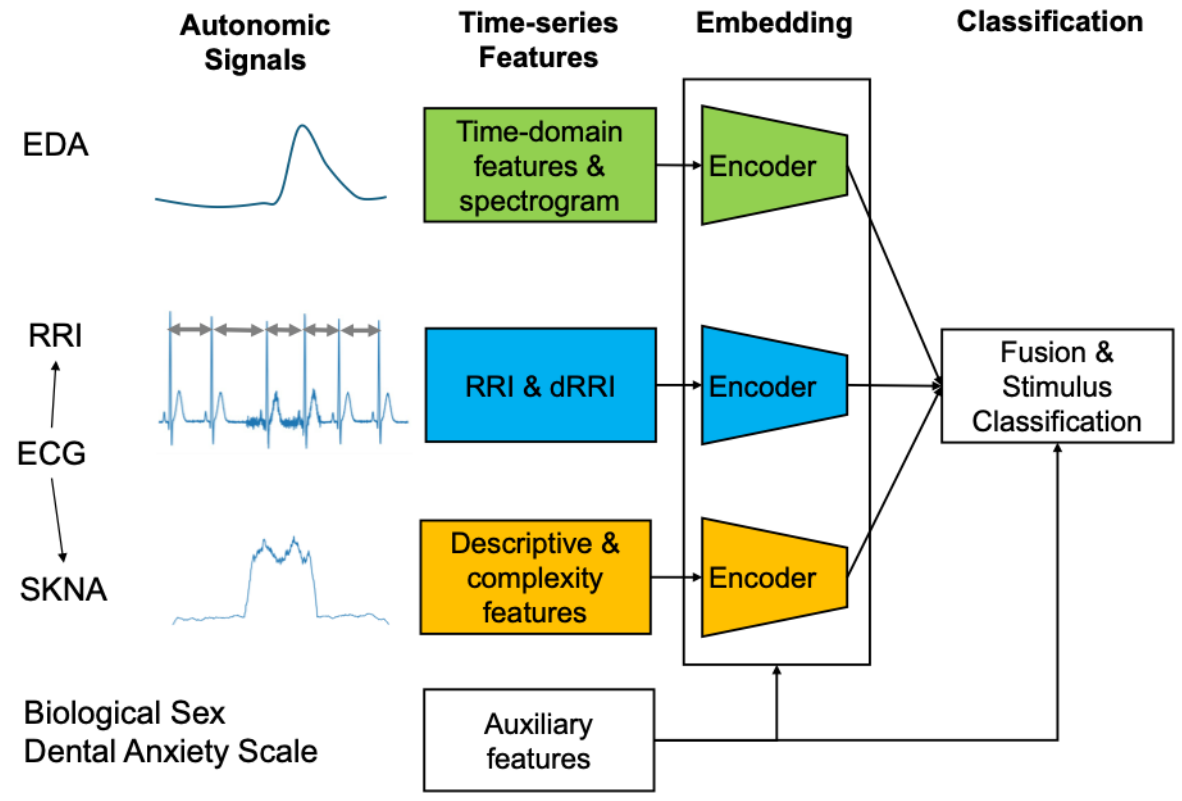

**Fig. 3.** Overview of proposed multimodal framework. Each modality uses a separate encoder with independent parameters.

1) **Segment size selection**

Each autonomic modality exhibits distinct temporal characteristics in response to stimulation. Cardiac responses, reflected in heart rate dynamics, are primarily mediated by adrenergic mechanisms (e.g., norepinephrine) and typically evolve more slowly than sudomotor activity. In contrast, EDA reflects sympathetic cholinergic activation of sweat glands via acetylcholine, resulting in more rapid responses that are closely linked to SNS activity. Consistent with these dynamics, prior studies have demonstrated effective use of EDA features within approximately 10 s windows [31].

Given the slower temporal response of cardiac dynamics, longer window lengths are required to capture meaningful variations in RRI. Accordingly, a 20 s window was used for RRI features, corresponding to the maximum duration available per stimulus.

SKNA, while also associated with sympathetic activation, captures neural activity at higher temporal resolution and has been shown in our prior work to respond more rapidly than EDA [53]. Therefore, a 6 s window was selected for SKNA feature extraction, which provides sufficient temporal coverage of TVSKNA dynamics (8.5 seconds, accounting for the 5-second temporal feature window) while maintaining compatibility with the downstream pooling strategy (e.g., divisibility by 4).

Table 1 summarizes the segment sizes for each modality, while Fig. 4 shows examples of modality-specific features.

TABLE I
SEGMENT SIZES FOR EACH MODALITY

| Mod. | Size | FS | Sequential Features |
|---|---|---|---|
| EDA | 10 s | 2 Hz | PhEDA, dPhEDA, PhdEDA, 8 STFT coefficients (n=11) |
| RRI | 20s | 4 Hz | RRI, dRRI (n=2) |
| SKNA | 6s | 2 Hz | TVSKNA descriptive and complexity features obtained from 5 s moving-windowing (n=11) |

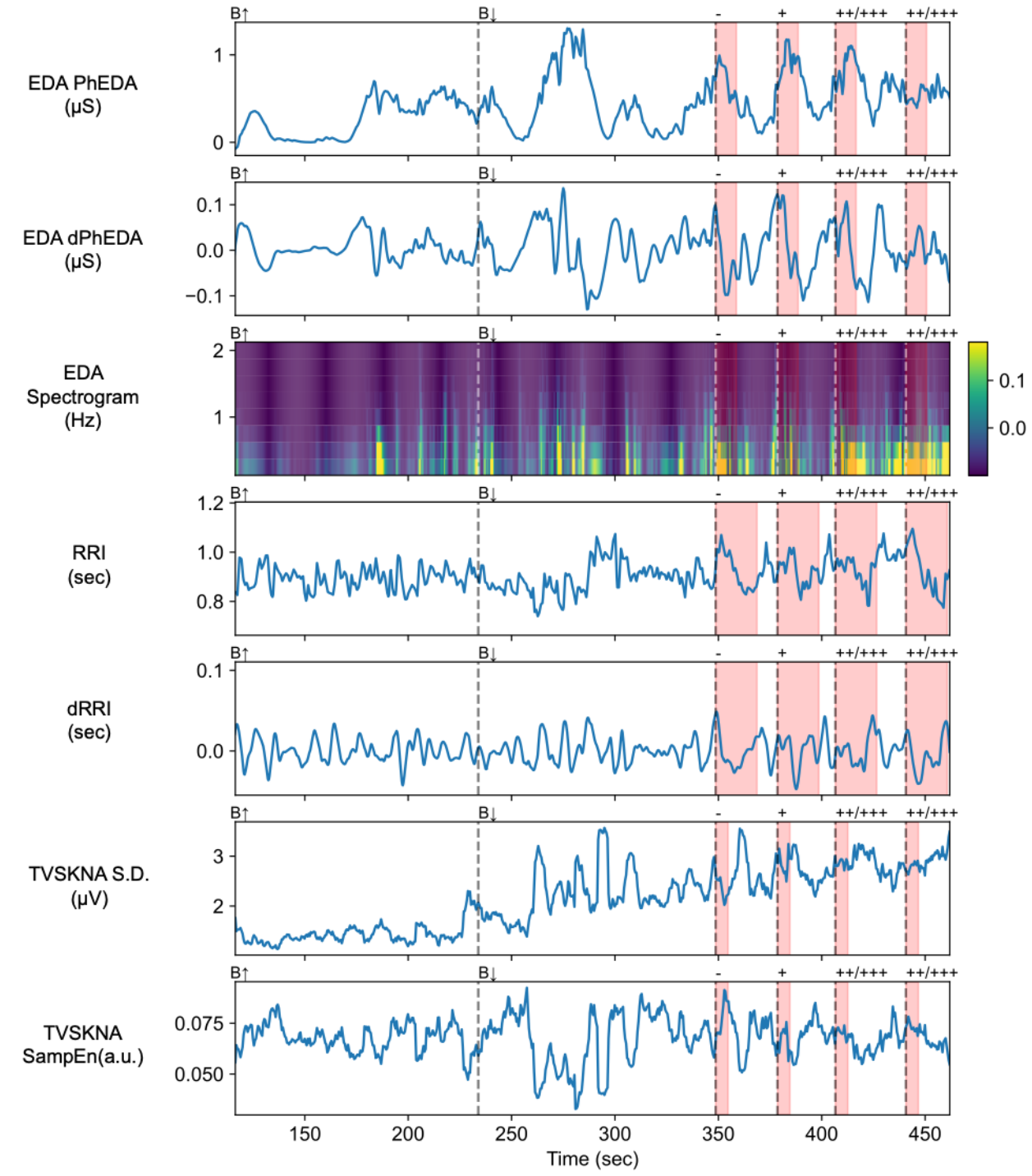

**Fig. 4.** Example of features. Red shades indicate window size.

2) **Dataset-level feature normalization**

To preserve the relative scale of each signal while ensuring balanced contributions across features within each modality, each signal channel was normalized using the global mean and standard deviation computed across all time points and all subjects in the training set. Normalization statistics were computed from the training subjects only and then applied to the held-out test subject.

$$\tilde{x}^{(i)}_{m,s}(t) = \frac{x^{(i)}_{m,s}(t) - \mu_{m,s}}{\sigma_{m,s} + \varepsilon}, \; m \in \{\text{EDA}, \text{RRI}, \text{SKNA}\}, \quad (1)$$

where $x^{(i)}_{m,s}(t)$ denotes the value of signal $s$ in modality $m$ at time $t$ for sample $i$, and $\tilde{x}^{(i)}_{m,s}(t)$ is normalized signal, and a small constant $\varepsilon = 1 \times 10^{-8}$ was added to the standard deviation to prevent division by zero. The parameters $\mu_{m,s}$ and $\sigma_{m,s}$ represent, respectively, the global mean and standard deviation of the corresponding signal, computed across all time points and all training subjects. Furthermore, additive Gaussian noise ($\sigma = 0.002$) was applied to the training input signals during training as a form of data augmentation. Non-temporal auxiliary variables, including sex and anxiety scores, were processed using the same normalization framework before model input.

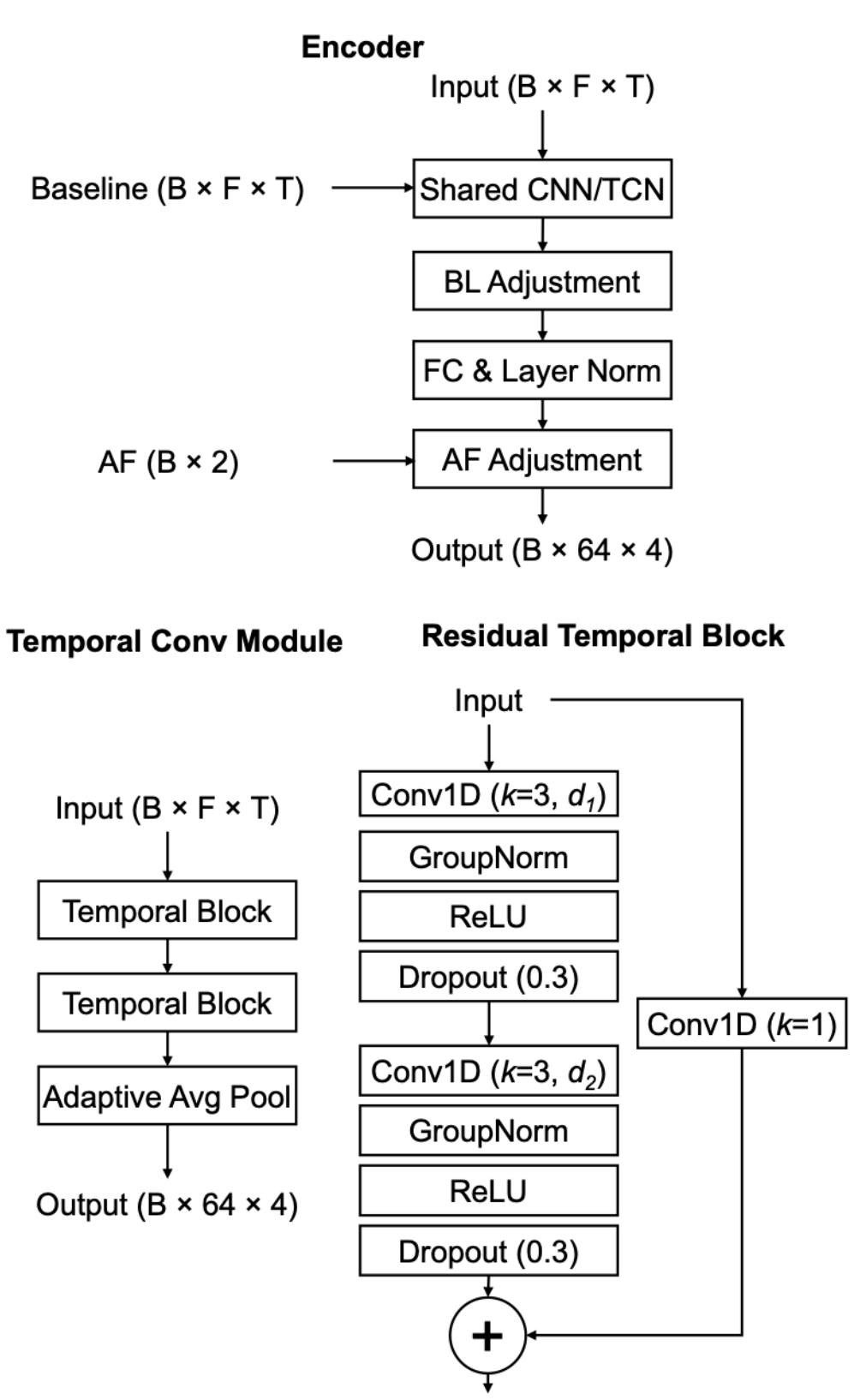

**Fig. 5.** Encoder structure. BL: baseline, AF: auxiliary features. B: batch size, F: feature numbers, T: sequential lengths, $d_1 = d_2 = 1$ for SKNA. $d_1 = 1$, $d_2 = 2$ for the others.

3) **Modality encoder structure**

a) Encoder

The encoder consists of four components: a temporal convolutional module, baseline adjustment, auxiliary-feature adjustment, and a projection layer for fusion. The temporal convolutional module can provide modality-specific temporal encoding, allowing heterogeneous autonomic signals with distinct temporal characteristics to be represented over appropriate time scales before multimodal fusion. The temporal convolutional module comprises two temporal blocks followed by adaptive average pooling, which reduces the sequence into four temporal bins while preserving temporal structure.

Each temporal block is implemented as a residual temporal block, consisting of two 1D convolutional layers (kernel size = 3), group normalization (4 groups) [54], and dropout (rate = 0.3) [55], with residual connections. The encoder produces a 64-dimensional embedding with four temporal bins via adaptive average pooling, preserving temporal structure.

The dilation configuration was determined based on the temporal characteristics of each modality. EDA and RRI signals have sufficiently long sequences (20 and 80 samples, respectively), allowing effective use of temporal convolutional networks (TCNs) to capture temporal dependencies [56]. In contrast, SKNA sequences are relatively short (12 samples), limiting the benefit of dilation-based temporal modeling. Therefore, TCN-based encoding was applied to EDA and RRI, while SKNA was processed using a residual CNN as the primary encoder. Fig. 5 presents the encoder architecture for each modality.

b) Baseline and auxiliary-feature adjustments

After the shared encoder, latent $Z_m$ and $Z_{baseline}$ are represented in the same latent space. Because baseline information serves as a reference, relative deviation from baseline should be emphasized rather than its absolute magnitude. To incorporate this effect, a baseline-adjusted latent representation is defined as follows:

$$\tilde{Z}_m = Z_m + tanh(\alpha) \times (Z_m - Z_{Baseline}), \quad m \in \{\text{EDA, RRI, SKNA}\} \quad (2)$$

where $\alpha$ is a learnable parameter to limit influence of deviation of each modality from baseline for each modality.

The projection layer for each modality then maps the encoded features to a unified latent representation using a fully connected layer followed by layer normalization [57], enabling modality fusion.

Subsequently, prior to attention-based fusion, auxiliary features are incorporated to gently modulate the normalized representations through scale and shift operations.

$$\tilde{\tilde{Z}}_m = \tilde{Z}_m \cdot (1 + a \cdot tanh(\gamma_m)) + b \cdot \beta_m, \quad m \in \{\text{EDA, RRI, SKNA}\} \quad (3)$$

where $\gamma_m$ and $\beta_m$ represent for learnable scale and shift parameters for each modality, respectively. The hyperbolic tangent maps $\gamma_m$ to the range [-1 to 1], ensuring stable modulation. The scaling factors, $a$ and $b$ were empirically set to 0.5 and 0.01, respectively. This design ensures that subject-level covariates provide contextual information without dominating modality-specific representations, thereby preserving the primary physiological signals while reducing the risk of overfitting given the limited dataset size.

4) **Fusion strategy**

The embedded latent representations from each modality are concatenated and subsequently processed using a multi-head attention mechanism with four heads to capture inter-modality relationships. The output of the attention layer is then concatenated with auxiliary features, followed by a fully connected layer for classification.

For binary classification, a sigmoid activation function is applied with binary cross-entropy loss, whereas for three-class classification, a Softmax activation function is used. Fig. 6. Illustrated the fusion strategy.

5) **Training parameters and validation scheme**

A leave-one-subject-out (LOSO) cross-validation scheme was employed. In each iteration, one subject was held out for testing, while the remaining subjects were used for training and validation. The training and validation sets were split in a 0.85/0.15 ratio using the stratified group K-fold scheme to ensure balanced class distributions while preventing subject overlap between sets. Among candidate splits, the partition with the smallest mean squared deviation from the ideal class distribution (i.e., number of samples divided by number of classes) was selected.

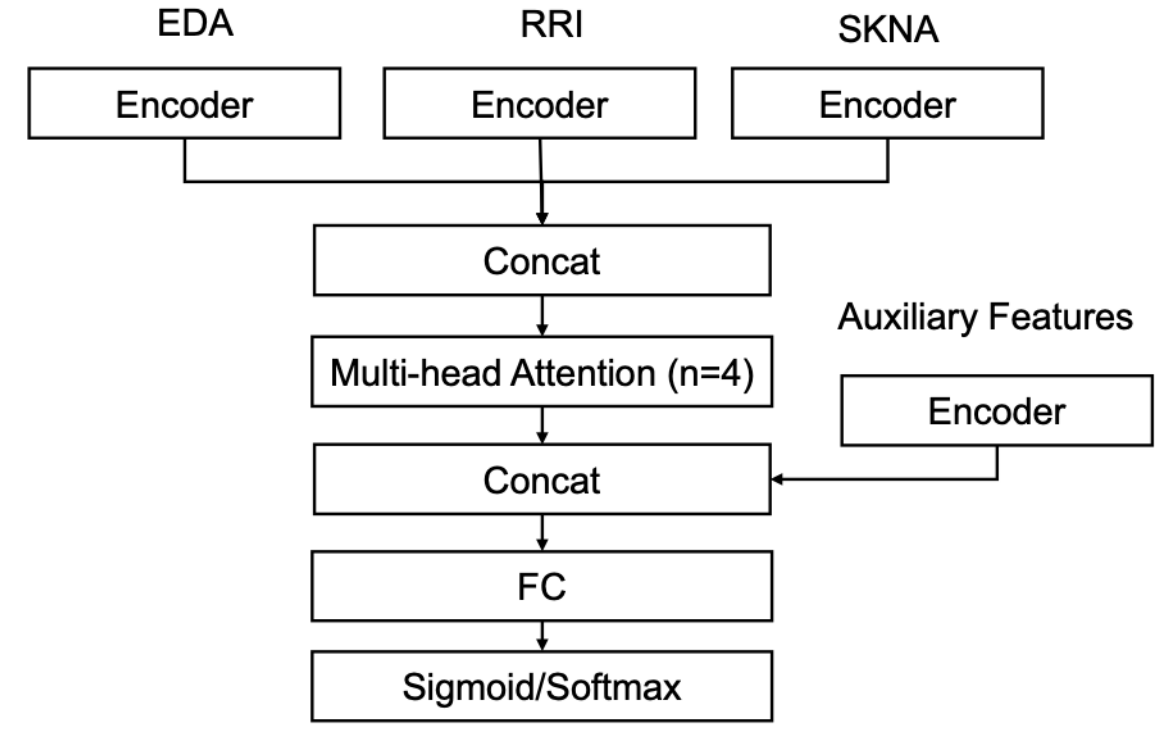

**Fig. 6.** Fusion strategy.

The validation set was used for early stopping (patience = 20) and learning rate reduction (patience = 10, reduction factor = 0.5) based on validation loss.

The model was trained with a batch size of 32 using the AdamW optimizer with a weight decay of $10^{-4}$ [58]. For binary classification, the learning rate was set to $5 \times 10^{-5}$ with 100 training epochs, while for thee-class classification, a learning rate of $1 \times 10^{-5}$ and 200 epochs were used.

For binary classification, binary cross-entropy with logits (BCEWithLogitsLoss) was used with a positive class weighting factor. For multi-class classification, categorical cross-entropy loss with class weights was applied.

6) **Compared Models**

Two deep learning-based baseline models were used for comparison with the proposed architecture. The first baseline model, CNN-Concat, used CNN encoders for all three modalities and directly concatenated the resulting modality-specific embeddings, instead of using attention-based fusion. The concatenated representation was then passed to a fully connected layer followed by a sigmoid or softmax output layer. The second baseline model, TCN-Concat, used the same modality-specific temporal encoders as the proposed model, including TCN encoders for the EDA and RRI modalities, but replaced attention-based fusion with direct concatenation. Both baseline models excluded the baseline-adjustment and auxiliary-feature-adjustment modules and used the same training and evaluation strategy as the proposed model.

Additionally, four conventional machine learning models were trained and compared, including RF, SVM with a linear kernel, LR, and eXtreme Gradient Boosting (XGB) [59]. The proposed model leverages convolutional networks to capture temporal structure, which is not explicitly modeled by conventional approaches. To ensure a fair comparison, all models were trained using the same set of time-series features derived from each modality.

For EDA and SKNA, descriptive and nonlinear features were extracted from the time-series signals. ApEn with m=2, r=0.20 was used instead of SampEn due to its numerical stability in short segments, as SampEn may yield undefined values. For RRI, given the 20 s window constraint, time-domain features were extracted, including mean, standard deviation, minimum, maximum, root mean square of successive differences (RMSSD), pNN20, mean absolute deviation of first-order differences (dRRI), standard deviation of dRRI and second-order differences, and linear trend slope.

The training pipeline consisted of feature standardization (except for RF and XGB), classifier-based feature selection, and hyperparameter optimization. Feature selection was performed using the same classifier as the final model. For RF and XGB, the maximum tree depth was limited to 4 during the feature selection classifier and to 16 for the final classifier to improve generalization. For all models, features with importance values greater than the mean were retained.

Feature selection and hyperparameter optimization were conducted using grid search with group-wise cross-validation (8 splits) to ensure subject independence, with the geometric mean of sensitivity and specificity used as the optimization metric. Candidate values for the regularization parameter (C) in SVM and LR were {0.01,0.1,1,10,100,1000}.

To account for class imbalance, balanced class weights were applied for SVM and LR, while a balanced subsampling strategy was used for RF. For XGB, class imbalance was addressed using class-balanced sample weights computed from inverse class frequencies in the training data. Finally, a LOSO cross-validation scheme was applied for evaluation.

### 7) **Evaluation metrics and statistical analyses**

For binary classification, balanced accuracy, sensitivity, specificity, and macro-averaged F1-score were computed. For three-class classification, balanced accuracy and macro-averaged F1-score were evaluated.

To further interpret the model, selected components across modalities were analyzed statistically, including attention weights and baseline adjustment coefficients (α). Attention weights were averaged across attention heads to obtain sample-level attention values for each modality. The baseline adjustment coefficient (α) is a scalar parameter for each fusion model; therefore, its value was directly analyzed without additional transformation (except for the bounded form tanh(α) used in the model).

Furthermore, auxiliary modulation parameters (β and γ) were examined to assess the influence of subject-level covariates on the learned representations. First, gradient-based sensitivity analysis was performed on the modulation pathways. Specifically, gradients of the latent representations with respect to auxiliary inputs were computed separately for the scaling (γ) and shifting (β) pathways. The absolute gradients were averaged across features and samples to obtain modality-wise importance scores:

$$importance\ of\ \gamma \text{ and } \beta = \left|\frac{\partial Z}{\partial AF}\right|, \quad (4)$$

where AF denotes auxiliary features. These parameter values were then averaged within each subject for each modality, and then normalized across modalities separately for each modulation pathway (γ and β). This normalization ensures that modality contributions sum to one for each subject, enabling comparison of relative importance.

To compare these attention weights and parameters, linear mixed-effects models were employed to account for within-subject correlations, except the baseline parameter $\alpha$. Post hoc comparisons were performed using estimated marginal means (EMMs) with Tukey correction [60]. Models were implemented using the LMER function from the lme4 package [61], with a random intercept for each participant:

$$importance\ of\ attention, \beta, \gamma \sim modality + (1|Participant), \quad (5)$$

$$\tanh(\alpha) \sim modality, \quad (6)$$

Finally, to evaluate the association between model performance and demographic variables (age, biological sex, race, and ethnicity), a generalized linear mixed-effect models (GLMM) with a binomial link function were employed as follows:

$$correct \sim DAS + Age + Biological\ Sex + Race + Ethnicity + (1|Participant) \quad (7)$$

where correct denotes whether a prediction was accurate (1) or not (0). Race was dichotomized into white and non-white groups due to limited sample sizes in minority categories, and ethnicity was categorized as Hispanic or non-Hispanic. A random intercept for participants was included to account for repeated measurements within subjects. Furthermore, odds ratios (ORs) were computed from the fitted model to quantify effect sizes, in addition to statistical significance testing [62]. An odds ratio of 1 indicates no association, values greater than 1 indicate increased odds, and values less than 1 indicate decreased odds of the outcome.

Statistical significance was defined as $p<.05$.

## IV. Results

Table 2 summarizes classification performance for both binary and multi-class tasks. Our framework achieved a balanced accuracy of 80.2% and an F1 score of 78.5% for binary classification (negative vs. positive responses), and 60.0% balanced accuracy with an F1 score of 58.8% for multi-class classification (negative vs. mild vs. intense responses).

TABLE II
CLASSIFICATION PERFORMANCE

| | | Binary classification (%) | | | | Multiclass classification (%) | |
|---|---|---|---|---|---|---|---|
| | | Balanced Acc | Sensitivity | Specificity | F1 | Balanced Acc | F1 |
| Proposed | All | 80.2 | 75.2 | 85.2 | 78.5 | 60 | 58.8 |
| Ablation study | -EDA | 66.5 | 61.4 | 71.6 | 64.8 | 51.8 | 50.2 |
| | -RRI | 71.3 | 64.1 | 78.4 | 69.2 | 51.7 | 49.9 |
| | -SKNA | 74.3 | 69.0 | 79.5 | 72.5 | 56.1 | 55.1 |
| | -BL | 79.1 | 75.2 | 83.0 | 77.6 | 59.9 | 58.6 |
| | -AF | 79.9 | 74.5 | 85.2 | 78.1 | 58.7 | 57.9 |
| Comparison | CNN–Concat | 79.5 | 73.8 | 85.2 | 77.7 | 55.1 | 53.9 |
| | TCN–Concat | 79.5 | 73.8 | 85.2 | 77.7 | 57.2 | 56.0 |
| | RF | 70.3 | 84.8 | 55.7 | 70.9 | 46.4 | 46.2 |
| | SVM | 72.0 | 69.0 | 75 | 70.6 | 51.1 | 51 |
| | LR | 72.3 | 69.7 | 75 | 71.0 | 52.8 | 52.2 |
| | XGB | 71.7 | 82.1 | 61.4 | 72.1 | 48.1 | 47.8 |

BL: baseline, AF: auxiliary features

Compared with conventional late-fusion tabular models, including RF, SVM, LR, and XGB, the proposed model consistently achieved higher performance. The best-performing conventional baseline achieved balanced accuracies of 72.3% and 52.8% for the binary and multiclass classification tasks, respectively. In contrast, the deep learning-based comparator models, CNN-Concat and TCN-Concat, achieved comparable performance to the proposed model for binary classification, with balanced accuracies of 79.5% for both models. For multiclass classification, however, these comparator models showed lower performance, achieving balanced accuracies of 55.1% and 57.2%, respectively.

Ablation analysis suggested that the EDA modality was a key contributor to binary classification, with performance dropping from 80.2% to 66.5% upon its removal. For multi-class classification (3-class), both EDA and RRI modalities were similarly important, with balanced accuracy decreasing from 60.0% to 51.8% and 51.7%, respectively, when each modality was removed.

Removing baseline and auxiliary features resulted in slight performance degradation (binary: 80.2% to 79.1% and 79.9%, respectively; 3-class: 60.0% to 59.9% and 58.7%, respectively), consistent with their intended role as complementary components with minimal but additive contributions.

Table 3 presents the confusion matrices for the proposed model. The model demonstrated reasonable performance in distinguishing dental stimulation from negative responses, as well as in separating extreme cases (i.e., negative vs. intense positive responses). In contrast, mild positive responses were more frequently misclassified between adjacent classes. This is likely due to variability in ANS responses, including influences from anxiety, individual sensitivity to cold stimulation, and inherent inter-subject variability.

Attention weight analysis further indicated that EDA contributed most prominently during attention fusion for binary classification (Fig. 7). For the multi-class task, both EDA and RRI exhibited comparable importance, with greater contributions than SKNA. These observations are consistent with our ablation study (Table 2). It should be noted that the linear mixed-effects models exhibited a singular fit, suggesting negligible between-subject variability in the estimated effects. This may indicate that the multi-head attention fusion operated consistently across participants.

TABLE III
CONFUSION MATRICES FOR BINARY (TOP) AND MULTI-CLASS (BOTTOM).

| | Neg | Pos | |
|---|---|---|---|
| Neg | 75 | 13 | |
| Pos | 36 | 109 | |
| | No (-) | Mild (+) | Intense (++) |
| No (-) | 64 | 14 | 10 |
| Mild (+) | 24 | 36 | 25 |
| Intense (++) | 12 | 9 | 39 |

Neg: no responses, Pos: positive responses.

Baseline parameters mildly amplified latent representations, as they are bounded between -1 and 1 (Fig. 8). Across both models, RRI exhibited significantly greater reliance on baseline adjustments compared to other modalities. This is likely because EDA and SKNA are primarily SNS-driven, monophasic signals that tend toward low activity at rest, whereas RRI reflects biphasic dynamics governed by both increases and decreases in heart rate.

Auxiliary feature analysis further indicated that RRI was significantly influenced by auxiliary features compared to other modalities (Fig. 9). This is consistent with prior findings that HRV is strongly modulated by biological sex and anxiety [49], [51], whereas EDA has shown more mixed or inconsistent associations with these factors [31], [63].

Table 4 presents odd ratios (ORs) for demographic variables from the fitted GLMMs. For binary classification, age was significantly associated with model performance (OR = 1.03, 95% CI [1.01, 1.07], p = 0.031), indicating that the odds of correct classification increased by approximately 3% for each additional year of age. A similar trend was observed for multi-class classification, although this did not reach statistical significance (OR = 1.02, 95% CI [1.00, 1.05], $p = 0.09$).

TABLE IV
ODD RATIOS FOR DEMOGRAPHIC VARIABLES

| | Binary classification | | Multi-class classification | |
|---|---|---|---|---|
| | Odd ratio | Sig. | Odd ratio | Sig. |
| Anxiety | 1.05 [0.95, 1.17] | *n.s.* | 0.99 [0.91, 1.09] | *n.s.* |
| Age | 1.03 [1.01, 1.07] | * (*p*=.031) | 1.02 [1.00, 1.05] | *n.s.* (*p*=.09) |
| Sex (Male) | 1.09 [0.53, 2.30] | *n.s.* | 1.71 [0.90, 3.49] | *n.s.* |
| Race (White) | 1.25 [0.58, 2.61] | *n.s.* | 1.84 [0.97, 3.77] | *n.s.* (*p*=.07) |
| Ethnicity (NH) | 1.04 [0.47, 2.19] | *n.s.* | 0.89 [0.45, 1.74] | *n.s.* |

For multi-class classification, race showed a trend toward higher performance in white subjects compared to non-white subjects, although this was not statistically significant (OR = 1.84, 95% CI [0.97, 3.77], p = 0.07). This trend may reflect the imbalance in the training dataset, which included a higher proportion of white participants (34 white vs. 15 non-white participants).

## V. DISCUSSION

Our study demonstrated the feasibility of using multimodal ANS signals for objective detection of dental pain. The proposed framework achieved reliable performance in distinguishing dental stimulation from negative responses, supporting the potential of non-invasive physiological monitoring as an adjunct tool for pulp vitality assessment.

Notably, our model demonstrated higher specificity (85.2%) than sensitivity (75.2%). This pattern is consistent with a prior EDA-based study [15], which reported 86.8% specificity and 76.2% sensitivity using the multi-layer perceptron approach. It should be noted that this comparison pertains only to the relative trend between sensitivity and specificity, as the prior study classified stimulation using a VAS ≥ 4 threshold versus painless segments without intermediate classes, whereas our model distinguishes negative versus positive responses using clinician-adjudicated labels, representing a more heterogeneous and inherently more challenging classification task. Given that EDA was the most influential modality in our model, this asymmetry may be attributed to the relative stability of EDA signals at baseline compared to stimulus-driven responses, which are more variable due to inter-subject differences. This finding is also consistent with our prior EDA-based classification study on high-intensity experimental pain [64], in which a random forest model achieved 78.9% sensitivity and 84.2% specificity. As the modalities used in this study are all derived from autonomic signals, improving sensitivity may require incorporating additional modalities beyond autonomic measurements.

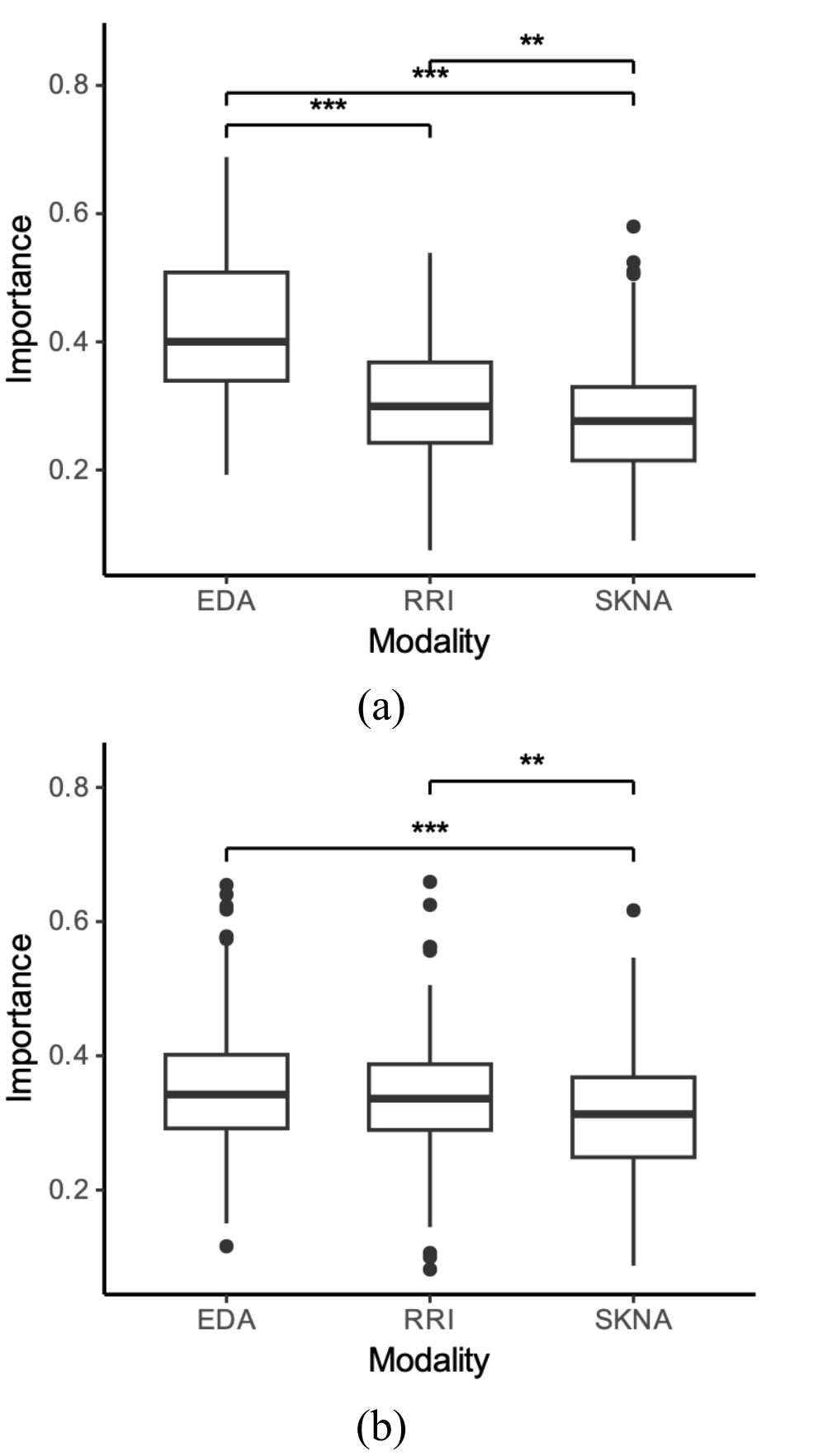

**Fig. 7.** Attention-weight-based importance comparisons: (a) binary, (b) multi class. ** $p < .01$, *** $p<.001$.

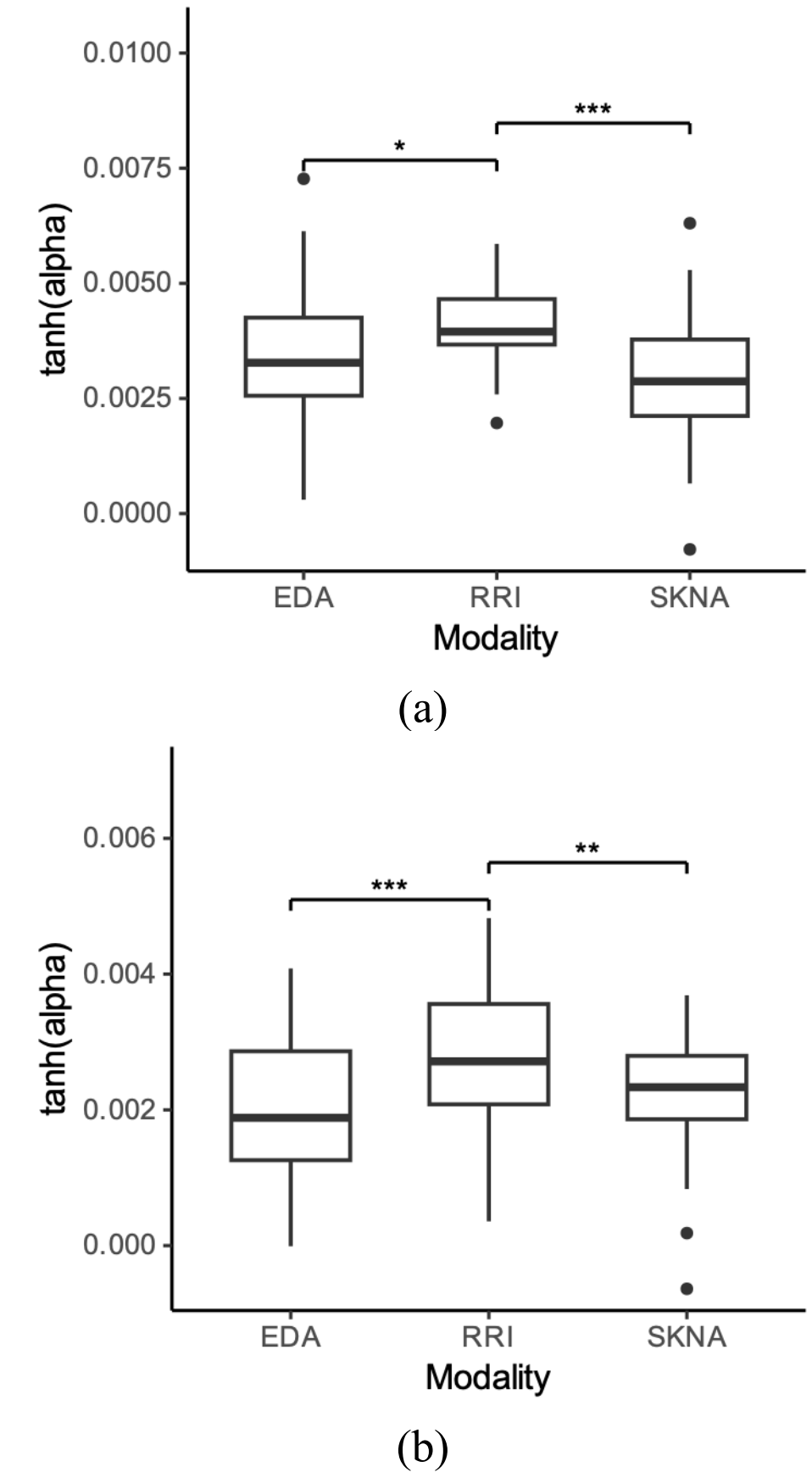

**Fig. 8.** Baseline parameter importance comparisons: (a) binary, (b) multi class. * $p<.05$, ** $p < .01$, *** $p<.001$.

Our modality contribution analysis indicated that EDA was the most influential modality, followed by RRI. Although SKNA contributed approximately a 5% improvement in balanced accuracy, this finding appears inconsistent with prior studies that have compared SKNA with other autonomic modalities. Previous work has shown that EDA can outperform HRV-derived indices during sympathetic activation tasks [65], [66]. However, to our knowledge, only a limited number of studies have directly compared SKNA with other modalities. For example, one study reported that SKNA outperformed EDA in classifying experimental sympathetic tasks, including thermal pain and the Valsalva maneuver [53]. This discrepancy may be attributed to the task-dependent nature of SKNA, as its performance can vary substantially across experimental conditions. In particular, clinical environments may introduce additional challenges, such as increased noise, motion artifacts, and anxiety-related confounders, which may reduce the reliability of SKNA-based features.

Interestingly, age emerged as an important demographic variable influencing model performance. Aging is typically associated with reduced autonomic function, particularly decreased parasympathetic activity and diminished dynamic modulation [67]. Given that pain responses are primarily mediated by sympathetic activation, the observed association between age and classification performance may reflect age-related reductions in parasympathetic activity and overall autonomic variability. This reduction in baseline variability may enhance the contrast between resting and stimulus-induced responses, thereby facilitating classification. However, further studies are needed to validate this interpretation.

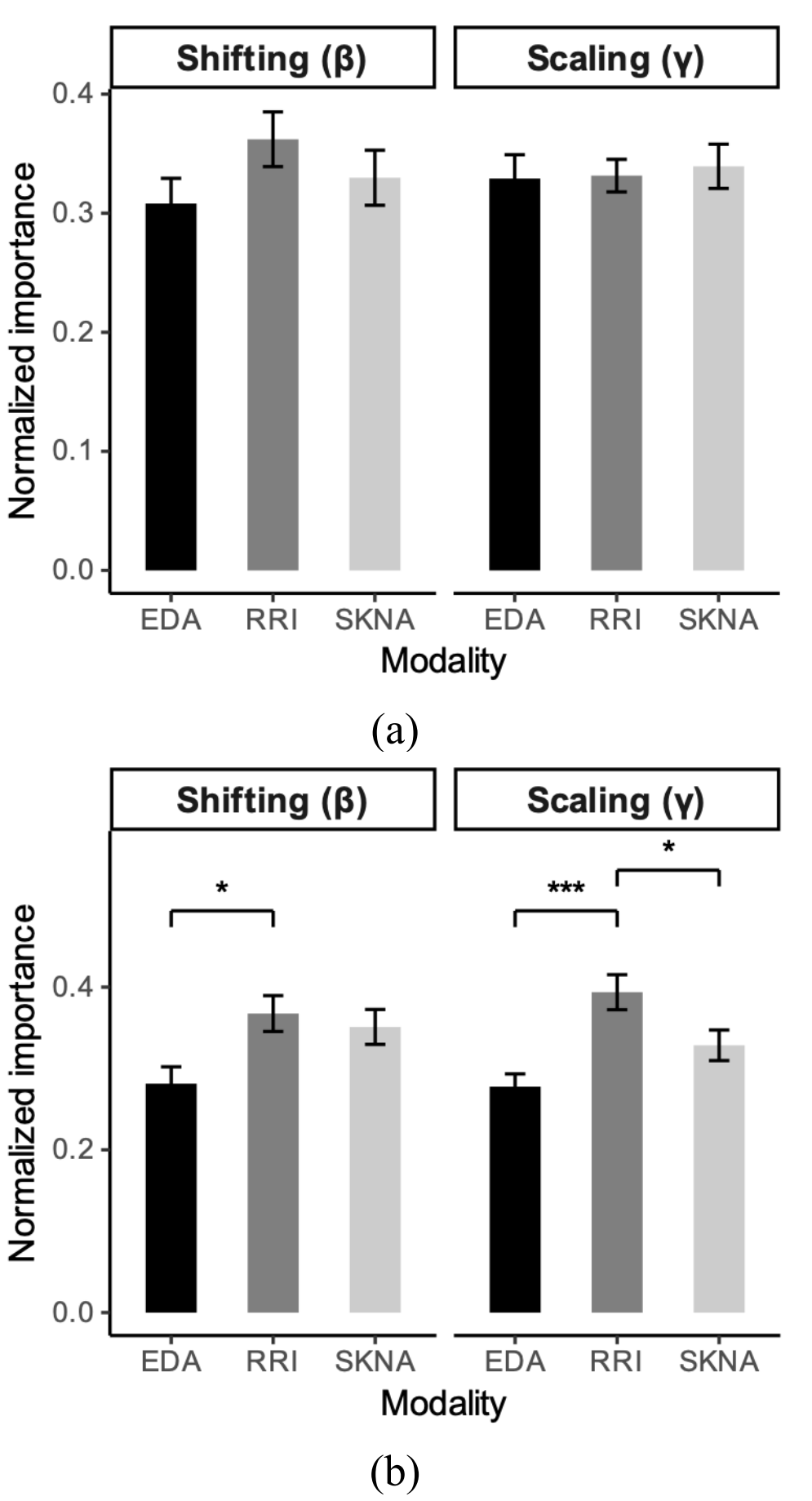

**Fig. 9.** Normalized importance for auxiliary feature parameters. (a) binary, (b) multi-class * $p$<.05, ** $p$ < .01, *** $p$<.001.

Our results demonstrate the feasibility of automatic dental pain detection in clinical settings using simple, non-invasive modalities. However, several limitations should be noted. The primary limitation of this study lies in the intermediate class within the multi-class classification task, which exhibited reduced separability from adjacent classes. Addressing this limitation may require more direct and objective measurements, such as modalities that capture pulp-specific signals or central nervous system activity. This is because ANS outputs reflect global physiological responses that are influenced by multiple confounding factors.

In addition, the dataset size remains limited. In particular, the model may be influenced by imbalance in racial representation, which may contribute to the weak trend observed in the multi-class classification results. Future work should incorporate larger and more diverse datasets, as well as additional auxiliary features that may influence ANS responses in dental clinical settings.

Motion and noise artifacts also present significant challenges. Although data in this study were collected under relatively controlled conditions and visually inspected for potential artifacts, real-world deployment will require automated artifact detection algorithms. While established methods exist for EDA and ECG signals [68], [69] , SKNA currently has only limited preliminary work [70]. Therefore, robust methods for automatic artifact detection and removal should be further developed and integrated into the proposed framework.

Despite these limitations, the proposed model demonstrated reliable performance in distinguishing dental stimulation from negative responses, supporting its potential as an objective tool for pulp vitality assessment.

## V. Conclusion

In this study, we proposed a multimodal framework for objective assessment of dental pain responses using non-invasive autonomic signals, including EDA, SKNA, and RRI. The proposed framework, designed with physiologically informed features and modality-specific configurations, demonstrated strong performance in differentiating negative from positive cases, while achieving reasonable performance for multi-class classification and outperforming conventional tabular feature-based approaches. In addition to classification performance, demographic analysis provided physiological insight, suggesting that age may influence model performance, whereas other factors showed limited effects.

Despite limitations related to dataset size, class separability,

and signal artifacts, these findings support the potential of non-invasive physiological monitoring as an objective and scalable tool for pulp vitality assessment in clinical settings. Future work should focus on expanding dataset diversity, integrating additional modalities, and developing robust artifact detection methods to enhance real-world applicability.

## Acknowledgment

The authors would like to thank Laura Doherty for her helpful discussions. During the preparation of this manuscript/study, the authors used ChatGPT 5 (OpenAI) for the purposes of grammar checking and phrasing improvements. The authors have reviewed and edited the output and take full responsibility for the content of this publication.